# Hierarchical Narrative Analysis: Unraveling Perceptions of Generative AI

Riona Matsuoka 1) Hiroki Matsumoto 2) Takahiro Yoshida 3) Tomohiro Watanabe 2) Ryoma Kondo 2),3) Ryohei Hisano 2),3)

1) Graduate Schools for Law and Politics, The University of Tokyo 2) Graduate School of Information Science and Technology, The University of Tokyo 3) The Canon Institute for Global Studies

***Abstract***: Written texts reflect an author's perspective, making the thorough analysis of literature a key research method in fields such as the humanities and social sciences. However, conventional text mining techniques like sentiment analysis and topic modeling are limited in their ability to capture the hierarchical narrative structures that reveal deeper argumentative patterns. To address this gap, we propose a method that leverages large language models (LLMs) to extract and organize these structures into a hierarchical framework. We validate this approach by analyzing public opinions on generative AI collected by Japan's Agency for Cultural Affairs, comparing the narratives of supporters and critics. Our analysis provides clearer visualization of the factors influencing divergent opinions on generative AI, offering deeper insights into the structures of agreement and disagreement.

## 1. Introduction

Written texts reflect the author’s perspective on various matters, making a thorough reading of the literature a well-established research method in many humanities and social science fields. For instance, legal scholars and professionals analyze court rulings to interpret norms shaped by judicial decisions. Similarly, policy makers incorporate insights from expert meetings and public feedback into their proposals. In Japan, following the recent revision of the Public Records Management Law enforcement ordinance and the transition to electronic public records management, there has been increasing demand for effective tools to analyze both academic and public documents. Consequently, quantitative analysis of narrative structures in texts has emerged as a significant focus across various disciplines [1-5].

The analysis of narrative structure becomes particularly engaging when significant differences among authors are identified. Whereas this type of analysis can be performed using traditional natural language processing techniques, such methods are often shallow. For example, sentiment analysis can assess the proportion of positive and negative opinions, and topic modeling [6] can extract keywords to highlight the topics each author focuses on. Furthermore, sentences containing causal expressions can be extracted and connected to summarize the causal structures recognized by the writer [7,8].

A more detailed investigation into narrative structures is presented in a series of studies summarized in [1]. In [1], narratives are defined according to [9] as “a non-random, connected sequence of events.” These studies use techniques such as dependency parsing and coreference resolution to extract key components, including the characters involved in the story, the events they participate in, and the temporal data indicating the chronological order of these events for further analysis. Though these methods are valuable, they do not account for the hierarchical nature of the writer’s narrative structure, making it difficult to grasp the central themes and the accumulation of facts leading to them. Furthermore, without properly organizing the hierarchical structure of the writer’s narrative, identifying the primary cognitive differences between writers becomes challenging.

In this paper, we propose a method for hierarchically aggregating the argumentative structures of writers by leveraging large language models (LLMs). The proposed method introduces a hierarchical structure template and uses LLMs to extract elements from the original text that align with this template, thus organizing the writer’s claims and the underlying perceptions that directly support them. Ultimately, this approach aims to visualize differences among writers through network analysis, using attributed information from documents (e.g., writer identity or ideological positions). Given that more complex tasks tend to result in higher error rates when relying solely on single queries to LLMs, the process is divided into three stages. To validate the effectiveness of this approach, we analyze public opinions released by the Agency for Cultural Affairs of Japan, comparing the narrative structures of those who are supportive of generative AI with those who are critical of it.

All queries and analyses were originally conducted in Japanese. Therefore, we have included the Japanese version of the manuscript in the appendix. For interested readers, we provide access to our repository, where all the data, code, and results from this paper are available.

## 2. Data

Whereas generative AI is innovative, it also carries the risk of causing social and ethical issues. The Subcommittee on Legal Systems of the Copyright Committee under the Agency for Cultural Affairs [10] sought public opinions on the relationship between AI and copyright, aiming to ensure fairness and transparency in administrative operations and to protect the rights and interests of the public. A total of 2,998 responses, along with respondent data, were made publicly available under the title “Public Comments on Perspectives Regarding AI and Copyright (Individual) Volumes 1–3” [10]. In this paper, we use these data to directly analyze public opinions on generative AI and copyright.

First, all opinions were classified into positive and negative categories. Figure 1 presents the results of polarity determination for each opinion, as conducted by “gpt-4-turbo-2024-04-09.” Specifically, the following prompt was used.

> The following text contains an opinion on the regulation of generative AI. Please evaluate the author's stance towards generative AI on a numerical scale, where a value closer to 1 indicates a positive stance, and a value closer to 0 indicates a negative stance.

As shown in Figure 1 and summarized in Table 1, which provides descriptive statistics regarding the text, there is a clear bias toward negative opinions, with fewer positive opinions observed. However, it is also evident that some individuals hold a positive view. In the subsequent analysis, opinions were classified into positive and negative categories based on a threshold of 0.5. This classification resulted in 277 positive opinions and 2,721 negative opinions. Though there may be debate over whether these data accurately reflect the polarity distribution of opinions on generative AI within society, or in other words, whether they represent a sample survey that is truly representative of public opinion, this paper will use these data to analyze the differences between positive and negative opinions on generative AI.

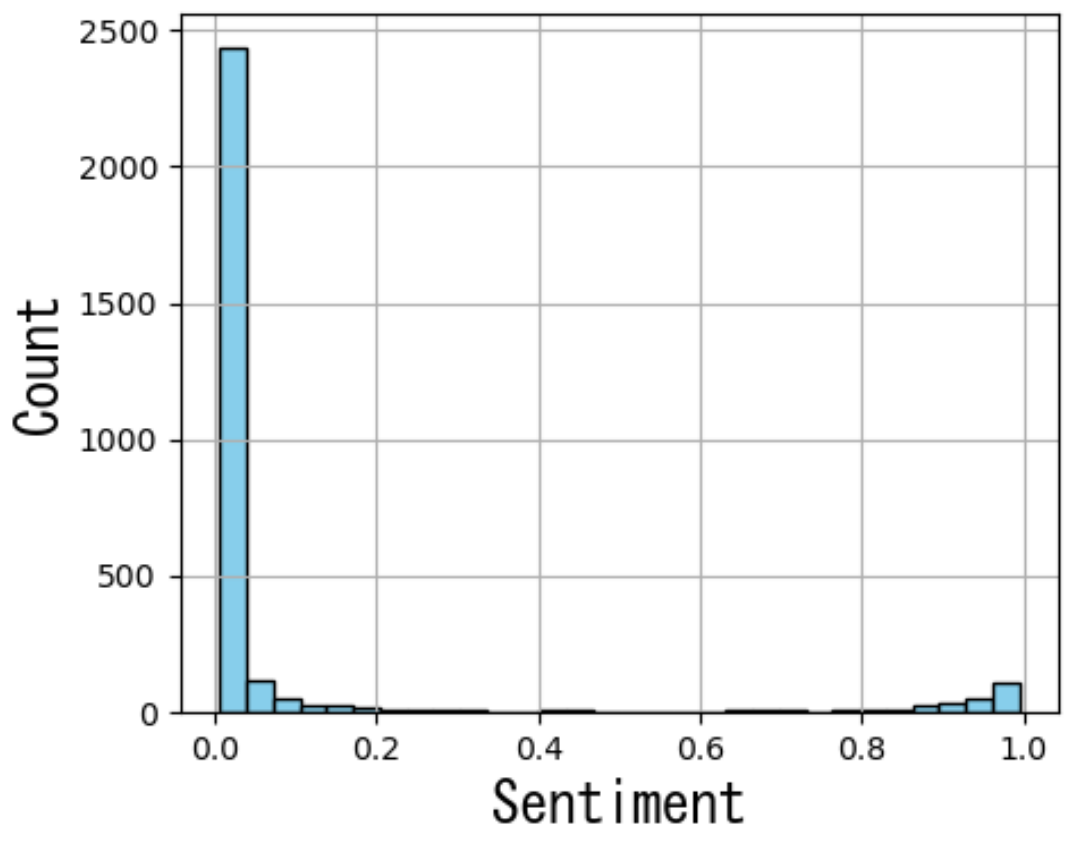

Figure 1. Sentiment Histogram

| | Mean | Std. | Min | Median | Max |
|---|---|---|---|---|---|
| **Character count** | 458.4 | 438.8 | 6 | 308 | 2015 |
| **Sentiment** | 0.11 | 0.26 | 0.01 | 0.01 | 0.99 |

Table 1: Descriptive Statistics of the Data

## 3. Proposed Methods

### (1) Hierarchical Structuring of Opinions

When analyzing the narrative structure of a text, simply instructing LLMs to do so without specific guidance may result in improper structuring. Therefore, opinions were structured with reference to Figure 2. This hierarchical diagram has two key points. First, the various elements discussed within the opinion are classified as the writer's perceptions, and after extracting the final main claims, such as requests or normative claims, the perception most directly related to these claims is selected as the primary perception. In reality, many opinions tend to be verbose. The core of the proposed method is to extract the significant perceptions from within this verbosity.

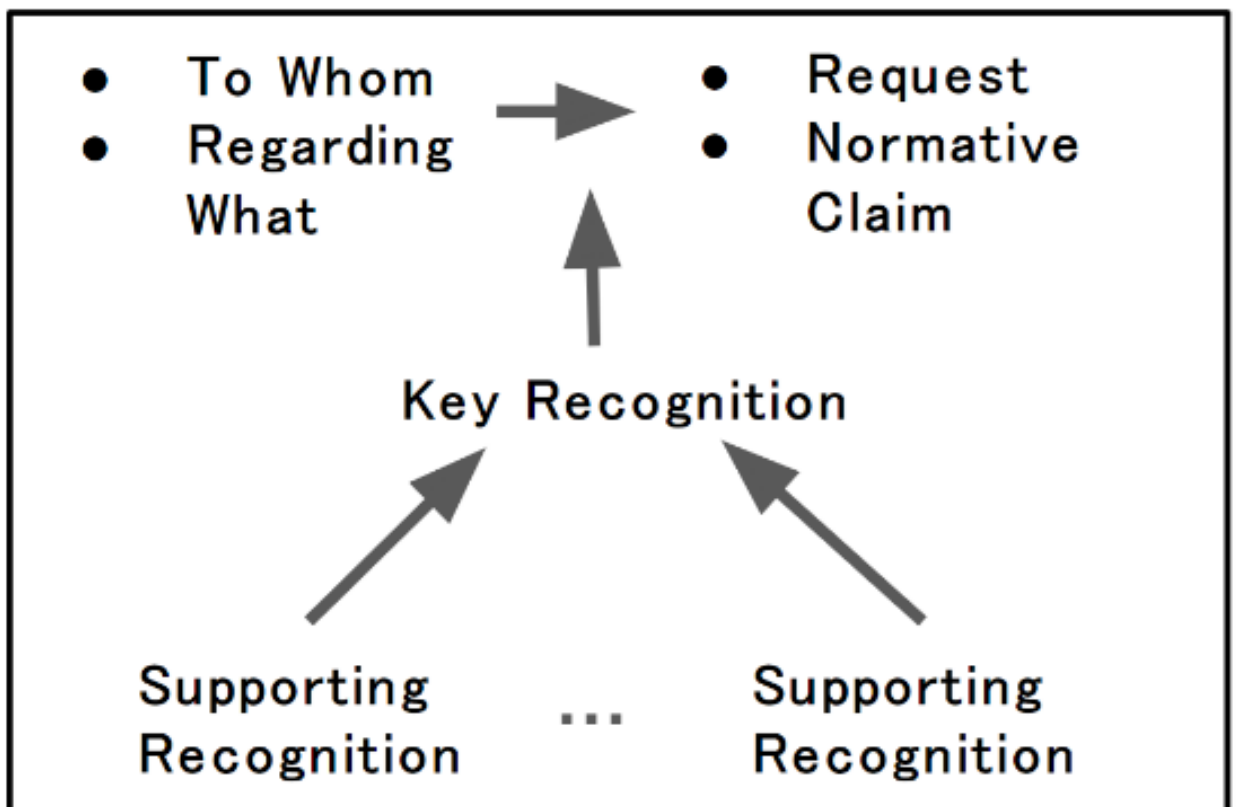

Figure 2: Hierarchical Structure

### (2) First Stage

In the first stage, keeping Figure 2 in mind, the sentences within the public comments were classified into the following four patterns:

**Pattern 1: Normative claims, expressed as "Who, regarding what, should be in what state."**
**Pattern 2: Requests, expressed as "Who wants whom to do what."**
**Pattern 3: Causal relationships, expressed as "What/who is causing what situation for whom."**
**Pattern 4: Perceptions, expressed as "Who perceives what in what way.**

It was observed that, when querying ChatGPT all at once, it tended to skip some sentences, even when they could be classified into the aforementioned patterns. To prevent such issues, the following adjustments were made:

1. Replaced pronouns and output rephrased expressions.
2. Instead of inputting all the opinions at once, extracted them in groups of five sentences.
3. Repeatedly emphasized, "Please use the original expressions as much as possible and avoid changing the wording unnecessarily."
4. Ensured that the edge list in the output did not contain any phrases that were not present in the original text, and verified that the occurrence of phrases not found in the original text was zero.

The specific prompt used is provided below. Due to space limitations, some parts have been omitted. The full prompt can be found in the publicly available code [17].

> The following text contains opinions regarding the regulation of generative AI. Please analyze it according to the following procedure:
> Identify the references of demonstrative pronouns such as "this," "that," and "those" used in the text, and output a replacement list for these pronouns.
> When creating the edge list, refer to this replacement list, and ensure that these demonstrative pronouns are replaced with their specific references.
> Extract information in the form of an edge list based on the following patterns. Almost all sentences can be classified into one of these patterns, so please categorize as many sentences as possible. Additionally, try to use the original expressions as much as possible, and avoid altering the wording unnecessarily. If the “who” part of Pattern 1, Pattern 2, or Pattern 4 refers to the author of the opinion, label it as “author.”
Pattern 1: Normative claims, expressed as “Who, regarding what, should be in what state.”
Pattern 2: Requests, expressed as “Who wants whom to do what.”
Pattern 3: Causal relationships, expressed as “What/who is causing what situation for whom.”
Pattern 4: Perceptions, expressed as “Who perceives what in what way.”
...Omitted for brevity...
Note: Pattern 4 can sometimes encompass Pattern 3. For example, Pattern 4 might be expressed as “Who perceives (what/who is causing what situation for whom).” In such cases, prioritize using Pattern 3. Ensure that all sentences are checked to determine if they fit into any of Patterns 1 through 4. The rephrased expressions included in the replacement list must be used in the edge list. Additionally, ensure that the edge list does not contain any phrases that are not present in the original text, and verify that the

An example of the output at this stage is as follows: from a total of 2,998 opinions, 5,637 instances of Pattern 1, 1,892 instances of Pattern 2, 8,527 instances of Pattern 3, and 6,227 instances of Pattern 4 were extracted.

“Replacement of Demonstrative Pronouns”:
“Here = ‘Concerns of those working in jobs equivalent to image creators’”
“Edge List”:
Pattern 3: (Generative AI, Creators and Performers, Causing concerns)
Pattern 3: (Excessive AI regulation, Creators, Posing a risk of significantly diminishing the benefits of AI technology)
Pattern 4: (Author, (Generative AI, Creators and Performers, Causing concerns), Perceives it as concerns regarding various issues)
Pattern 4: (Author, (Excessive AI regulation, Creators, Posing a risk of significantly diminishing the benefits of AI technology), Perceives it as excessive regulation)
Pattern 3: (Rising prices and declining birthrate, Around creators, Causing a shortage of human resources)

### (3) Second Stage

Patterns 1 through 4 do not sufficiently analyze the requests and normative requirements outlined in Figure 2. Therefore, in the second stage, additional patterns were extracted to identify the perceptions directly related to these requests and normative requirements, as well as to derive the indirect perceptions that support them. This step builds upon the four basic patterns by incorporating the following framework.

・**Pattern 5: Identifying the relationship when a causal relationship (Pattern 3) or perception (Pattern 4) influences other elements, such as normative claims (Pattern 1) or requests (Pattern 2).**

For the extraction of Pattern 5, instructions were given to input both the original text and the patterns (Pattern 1 through Pattern 4) extracted in the first stage for each opinion, and then extract Pattern 5. In the second stage, the following additional measures were implemented in addition to those from the first stage:

1. Ensure that no output is generated in a format that exceeds the specified edge list structure, such as tuples with more elements than allowed.
2. For all instances of Pattern 1 and Pattern 2, ensure that corresponding relationships are identified, and if there are no corresponding normative claims (Pattern 1) or requests (Pattern 2), instruct the model to return an empty edge list.

The specific prompt used was as follows:

Next, based on the original text and the previously extracted results for Patterns 1 through 4, please follow the steps below for analysis:
First, clarify what demonstrative pronouns such as “this,” “that,” and “those” refer to within the text... [omitted]... Then, extract information in the form of an edge list based on the following patterns. Ensure that you use the original expressions as much as possible and avoid altering the wording in the text. There is no need to extract Patterns 1 through 4 again; this time, focus exclusively on extracting Pattern 5.
Pattern 5: Identifying the relationship when a causal relationship (Pattern 3) or perception (Pattern 4) influences other elements, such as normative claims (Pattern 1) or requests (Pattern 2).... [omitted]...
For all instances of Pattern 1 and Pattern 2, verify their correspondence with the previously extracted Patterns 1 through 4 and determine which condition of Pattern 5 applies. If the “who” part in the text seems to refer to the author of the opinion, label it as “author.” Ensure that the rephrased expressions included in the replacement list are used in the edge list. Additionally, confirm that the output edge list does not contain any phrases that are not present in the original text, and guarantee that the occurrence of phrases not found in the original text is zero. Furthermore, strictly adhere to the specified edge list format, and do not output tuples with more elements than allowed. If there are no normative claims (Pattern 1) or requests (Pattern 2), return an empty edge list. Extracted patterns should be outputted without altering a single word from the original text. The most important instruction is not to modify the original text without permission.... [omitted]...

The output results of the second stage were as follows: through this step, 6,054 instances of Pattern 5 were extracted.

“Replacement of Demonstrative Pronouns”:
“This possibility” = “The possibility of increased productivity through generative AI”
“That opportunity” = “The opportunity for increased productivity through generative AI”
“Edge List”:
Pattern 5: (Author, (Prices, Creators’ compensation, Causing a situation where compensation becomes relatively lower), (Staff and workload management, Being required to maintain the quality of outputs))
Pattern 5: (Author, (Stagnation of generative AI technology as a whole, Domestic cultural activities, Causing a significant setback), (Creators and society as a whole, Wanting to move forward while balancing the protection of individual rights and the development of AI technology))

**(4) Third Stage**

In the form depicted in the diagram, it is difficult to grasp the relationships between each element at a glance, making analysis challenging. Therefore, ChatGPT was used again to convert the output data into sentences. Specifically, instructions were given to summarize each element of Pattern 5 into sentences of the form “Because A is B, C should be D” or “Because A is B, I want E to do F,” with each element being condensed to within 10 characters.

The specific prompt used is as follows.

The sentence provided represents either a request in the form "Since statement 1 is true, I would like statement 2 to happen" or a normative claim in the form "Since statement 1 is true, statement 2 should be the case."
The sentence can be summarized in one of two formats, depending on the nature of statement 2:
If statement 2 is normative, the summary should follow the format: “Since A is B, C should be D.” Here, “A is B” is extracted from statement 1, and “C should be D” or “C ought to do D” is extracted from statement 2.
If statement 2 is a request, the summary should take the form: “Since A is B, please do F to E.” In this case, “A is B” comes from statement 1, while “please do F to E” comes from statement 2.
Each component (A through F) should be no longer than 10 characters, excluding conjunctions and particles. The final summary should be provided accordingly: if statement 2 is normative, mark E and F as “NA”; if it is a request, mark C and D as “NA.”
Ensure that the edge list contains only phrases from the original sentence, with zero occurrences of phrases not present in the original text. (The text continues...)

Table 2 summarizes the results.

| type | A | B | C | D | E | F |
|---|---|---|---|---|---|---|
| Normative Claim | The State of Image Generation AI Without Requiring Permission | It significantly undermines the equality of benefits and opportunities for creators, copyright holders, and authors. | A State Where Image-Generating AI Does Not Require Permission | Non-Acceptance | NA | NA |
| Request or Demand | Generative AI | Economic Decline in Japan | NA | NA | Japanese Government | Help Japan |

**Table 2: Results After Stage 3**

**(5) Network Visualization**

The extracted output results, ranging from A to F, were transformed into vector embeddings using the Universal Sentence Encoder [11]. Next, clustering was performed using DBSCAN [12], grouping items with similar semantic content. The cluster numbers were denoted as A_c, B_c, C_c, D_c, E_c, and F_c. The representative expression for each cluster was chosen based on the central element within that cluster.

Subsequently, the connections A_c → B_c, B_c → C_c, C_c → D_c and A_c → B_c, B_c → E_c, and E_c → F_c were divided into positive and negative opinions and visualized as a directed network. In this network, the edges correspond to three types: recognition (“Because ◯ is □”), normative claims (“◯ should be □”), and requests (“Please do □ to ◯”).

For the overall network, the focus was on edges with a degree of two or more. However, for positive connections, due to their limited number, edges with a degree of one were also included. In addition, the positions of the nodes were arranged so that identical nodes, regardless of whether they were in the positive or negative group, would be drawn in the same location. The layout was generated using ForceAtlas2 [13], implemented in Gephi [14]. The color of each node represents the community detected through modularity maximization [15]. This community structure was calculated based on the union of the positive and negative networks.

**4. Results**

The networks of the negative group and the positive group obtained through the aforementioned process are shown in Figures 4 and 5, respectively. As previously mentioned, the common nodes in both networks are displayed in the same positions. Based on this information, the rough correspondence between positions is indicated by the numbers (1) through (7). The characteristics of each of these positional groupings will be discussed in detail.

First, at position (1), the central node is “AI-generated contents.” Whereas the negative group includes some opinions near position (1), suggesting that AI use should “not be overly restricted,” there is strong and widespread opposition toward “AI-generated contents.” This opposition includes statements such as “not acceptable,”

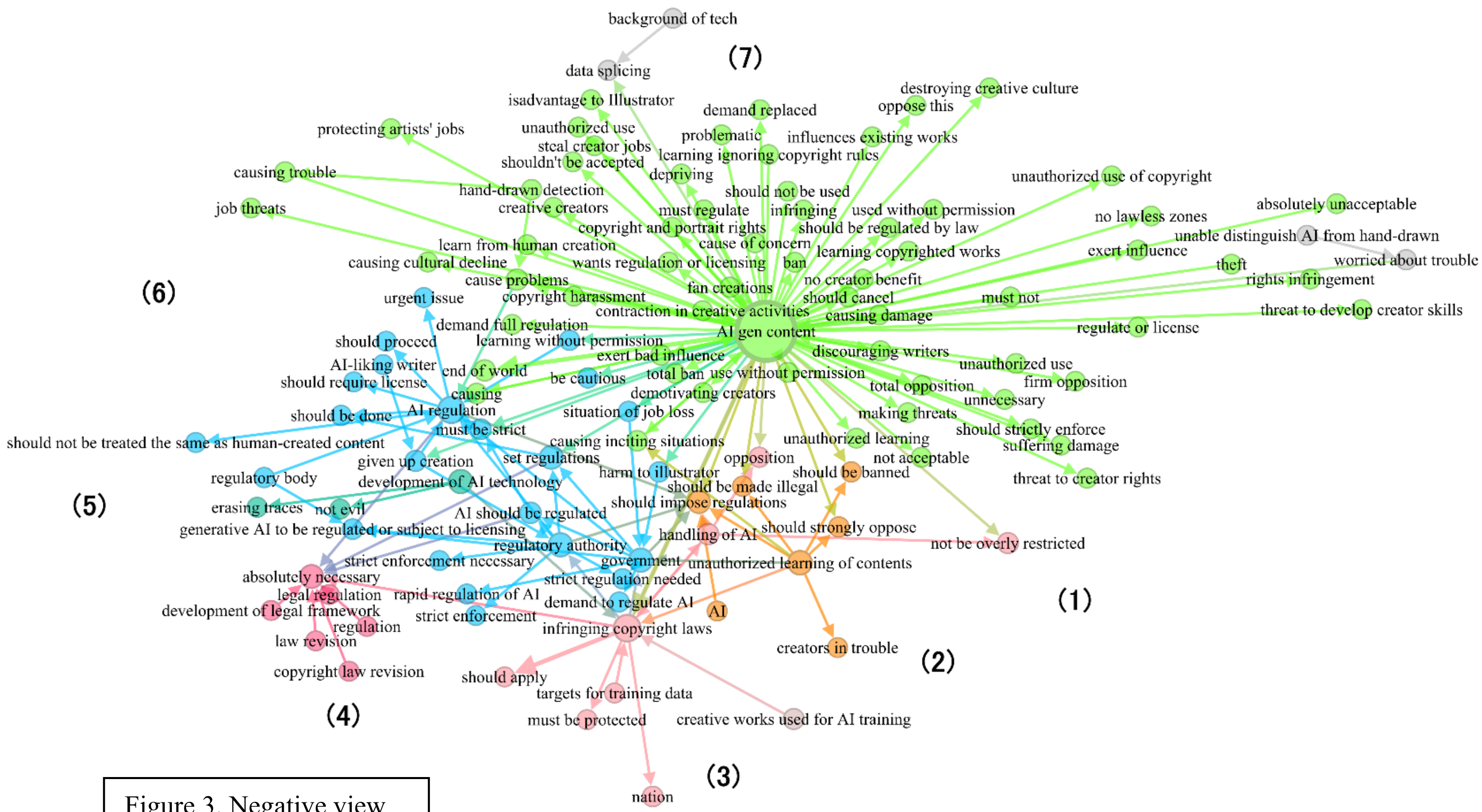

Figure 3. Negative view

"making threats," "no creator benefit," "cause problems," "theft," and "destroying creative culture." These views are prevalent throughout the network, making "AI-generated contents" the node with the highest degree in the network. In contrast, the positive group centers around the belief that "AI research" and "learning copyrighted materials" should "not be overly restricted." Although there is an acknowledgment that "unauthorized learning of contents" "lacks respect" for creators, more lenient opinions are noticeable regarding "AI-generated contents," such as "should not be stopped" and "copyright should be recognized." In other words, strong opposition similar to that in the negative group is absent.

Next, at position (2), the node that highlights the divergence in perception between the positive and negative groups is "unauthorized learning of contents" by AI. The negative group predominantly expresses concerns regarding "unauthorized learning of contents," with views such as "creators in trouble" and "should strongly oppose." In contrast, the positive group argues that "unauthorized learning of contents" by AI "should be fully accepted" and that restrictions related to "restriction by copyright" "should not hinder development" in the field of "computer science." Overall, positive opinions toward new technologies are more prominent within this group.

At position (3), the central node is "creative works used for AI training." The negative group strongly criticizes the use of "creative works used for AI training," arguing that it is "infringing copyright laws." In contrast, the positive group adopts a more engineer-oriented perspective, asserting that we "should not set" a "limit" on the "use of training data" concerning "creative works used for AI training." It is noteworthy that these opinions tend to align more with the engineering side rather than the creators' viewpoint. Although public comments do not include attribute information unless explicitly stated by the writer, it is interesting to note that certain attributes can be inferred to some extent by analyzing the narrative structure.

Positions (4) and (5) primarily focus on regulatory issues. The negative side strongly argues that "AI should be regulated" due to "violations of copyright" and calls for stricter regulations from the "regulatory authority." They express firm opinions, advocating for "generative AI to be regulated or subject to licensing" and demanding "strict enforcement." If we over read this statement and interpret a licensing system as a special permission-based system, it essentially means that citizens do not inherently have the right to freely engage in the relevant business, as the state holds a monopoly over it. Therefore, this might be seen as quite a strong position. In contrast, the positive side, which supports generative AI users, requests that "regulatory authorities" "provide specific examples and raise awareness," rather than imposing strict regulations.

Regarding the "development of AI technology" at position (5), the positive side strongly associates it with economic growth, stating that it is "synonymous with industrial development." In addition, nearby discussions emphasize the link between "promotion by foreign countries" and the "dangers of technological disparity." An interesting opinion from the positive side suggests that creators should also take the development of technology into account, stressing that "warm-up" activities are "absolutely necessary" to prevent the "development of AI technology" from leading to the "stagnation of creative activities." In contrast, the negative side connects the "development of AI technology" with concerns about "erasing traces." Upon reviewing the original opinions, this concern refers to the possibility that AI technology could erase traces that form the basis for copyright law, raising fears that further advancements in AI technology could render the current legal system ineffective.

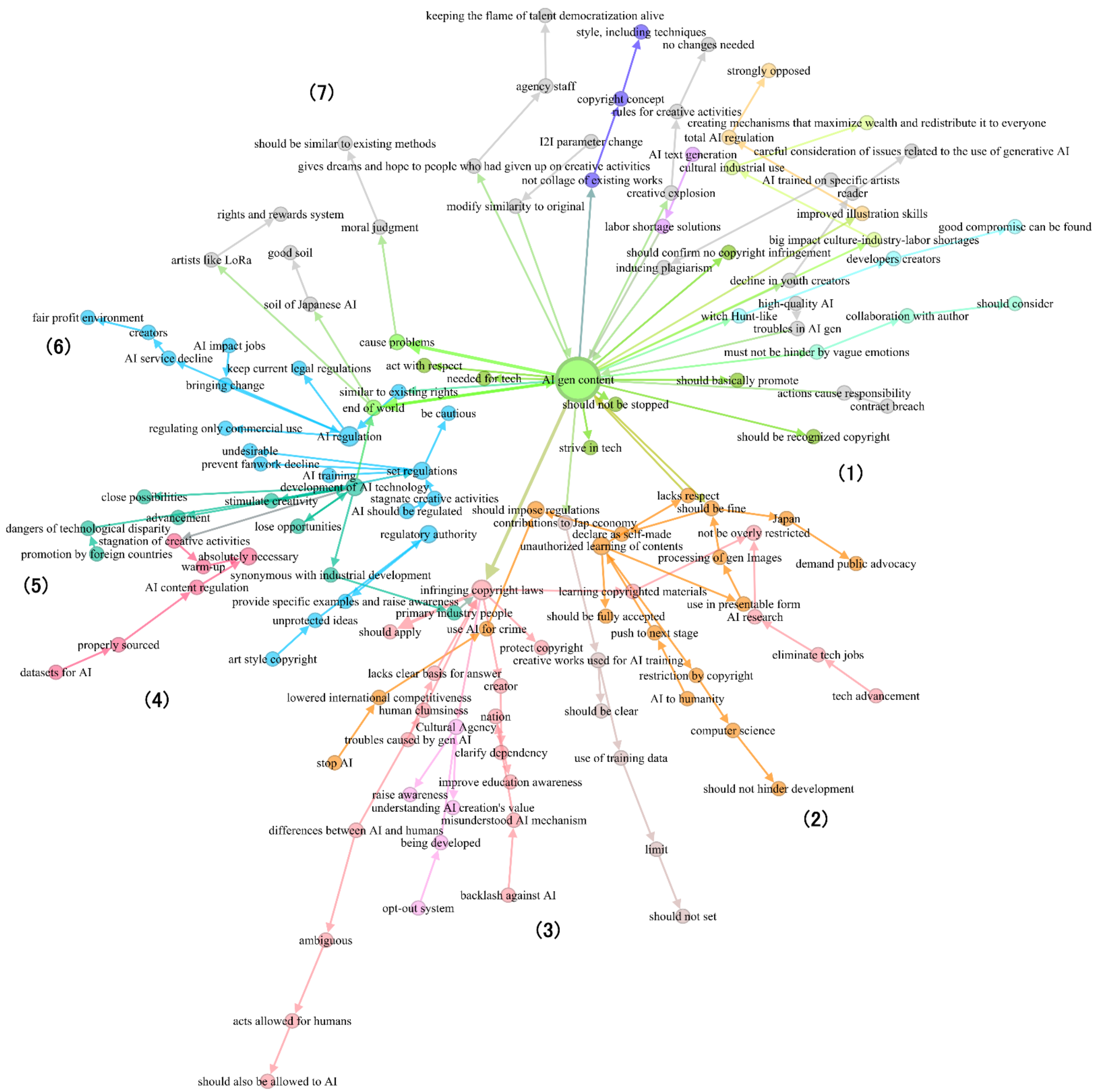

Figure 4. Positive view

Position (6) also centers on regulatory issues. The positive group generally advocates for maintaining the status quo regarding "AI regulation," expressing opinions such as "keep current regulations" or "regulate only commercial use." In contrast, the negative group adopts a more proactive stance, arguing that "AI regulation" "should require a license," "must be strict," and that AI-generated content "should not be treated the same as human-created content." This proactive approach is particularly noteworthy. Japan's copyright law was amended four times between 2018 and 2021. Notably, the 2018 amendment allowed the use of copyrighted works without permission for research and technological development, facilitating AI training (Articles 30-4, 47-4, and 47-5). Though it is unclear whether these amendments influenced the positive group's stance, their preference for maintaining the status quo may reflect an understanding of the legal system's support for machine learning applications.

Finally, at position (7), the negative group, similar to position (1), strongly opposes "AI-generated contents." This opposition includes statements such as "destroying creative culture," "theft," and "threat to developing creator skills." The widespread critical sentiment toward "AI-generated contents," which is the node with the highest degree, extends directly to this position as well. In contrast, the positive group expresses optimism about effectively utilizing the new technology. They advocate for "keeping the flame of talent democratization alive," emphasize the importance of "creating mechanisms that maximize wealth and redistribute it to everyone," and call for "careful consideration of issues related to the use of generative AI." They also highlight how generative AI "gives dreams and hope to people who had given up on creative activities" and express the hope that a "good compromise can be found."

| No. | Central node | Positive | Negative |
|---|---|---|---|
| (1) | "AI-generated contents" and "should not be overly restricted" | The positive group centers around the belief that "AI research" and "learning copyrighted materials" should "not be overly restricted." They advocate for "public engagement" with "AI," recognizing that unauthorized learning "lacks respect" for creators. However, more lenient views are expressed regarding "AI-generated contents," such as "should not be stopped" and "copyright should be recognized." Strong opposition, like that seen in the negative group, is absent. | Though there is an opinion that AI use should "not be overly restricted," the negative group expresses strong and widespread opposition toward "AI-generated contents." This includes statements like "not acceptable," "making threats," "no creator benefit," "cause problems," "theft," and "destroying creative culture." These views dominate the network, making "AI-generated contents" the node with the highest degree in the network. |
| (2) | "Unauthorized learning of contents" | The positive group argues that "unauthorized learning of contents" by AI "should be fully accepted." They also emphasize that restrictions related to "copyright" should not "hinder the development" of "computer science." Overall, positive opinions toward new technologies are more prominent in this group. | The negative group expresses significant concerns about "unauthorized learning of contents," with opinions such as "creators in trouble" and the belief that it "should be strongly opposed." |
| (3) | "Creative works used for AI training" | The positive group, taking a more engineer-oriented perspective, argues that there "should not be" restrictions on the "use of training data" concerning "creative works used for AI training." This viewpoint tends to align more with engineers rather than creators. | The negative group strongly criticizes the use of "creative works used for AI training," viewing it as "infringing copyright laws." |
| (4) | "Regulatory authorities" | The positive side, aligned with generative AI users, requests that "regulatory authorities" "provide specific examples and raise awareness," rather than imposing strict regulations. | The negative side advocates for stricter regulations, urging "regulatory authorities" to "regulate or license generative AI" due to "violations of copyright" and demands "strict enforcement." |
| (5) | "Advancement of AI technology" | The positive side strongly associates the "advancement of AI technology" with economic growth, describing it as "virtually synonymous with industrial development." They also highlight the "promotion by foreign countries" and the potential "dangers of technological disparity." In addition, there is a call for creators to engage in "warm-up" activities to prevent the "stagnation of creative activities." | The negative side links the "advancement of AI technology" to concerns about "erasing traces." This refers to the fear that AI could erase traces that underpin copyright law, potentially rendering the current legal system ineffective as AI technology progresses. |
| (6) | "Regulation of AI-generated content" | The positive group generally supports maintaining the status quo, with opinions such as "limiting regulation to the current legal framework" or "regulating only commercial use." This stance may reflect an understanding of the legal system's support for machine learning applications. | The negative group advocates for stricter regulation, expressing views that "AI regulation should require a license," "must be strict," and that AI-generated content "should not be treated the same as human-created content." This proactive approach emphasizes a stronger regulatory framework. |
| (7) | "AI-generated contents" | The positive group expresses optimism about the potential of generative AI. They advocate for "keeping the flame of talent democratization alive," emphasize "creating mechanisms that maximize wealth and redistribute it to everyone," and call for "careful consideration of issues related to generative AI." They also highlight how generative AI "gives dreams and hope to people who had given up on creative activities" and express the hope that a "good compromise can be found." | Similar to position (1), the negative group strongly opposes "AI-generated contents," with concerns such as "destroying creative culture," "theft," and "threat to developing creator skills." The critical sentiment toward AI-generated content, which is the node with the highest degree, extends directly to this position as well. |

Table 3. Summary of the two views

In conclusion, by consciously extracting the main arguments from public comments with attention to the hierarchical structure of opinions, we were able to compare and analyze the differences in the narrative structures between the negative and positive groups regarding generative AI. For clarity, a summary of the differences between the two groups is presented in Table 3.

## 5. Conclusion

In this paper, we proposed a method that utilizes Large Language Models (LLMs) to hierarchically organize the argumentative structure of a text and visualize the differences among speakers through network analysis using sentence attribute information. Unlike narrative structures that focus on elements such as people, relationships, and time [1], the proposed method is considered suitable for documents with clearly defined opinion structures, such as public comments or legal judgments. To demonstrate the effectiveness of the proposed method, we conducted an analysis using the results of public comments on generative AI published by the Agency for Cultural Affairs. First, we determined whether each opinion was positive or negative by assessing the overall polarity of the comments. We then conducted an analysis focusing on the differences in narrative structures between the positive and negative opinions. As a result, compared with conventional methods, we were able to more clearly visualize the factors that lead to differences in opinions on various issues.

As future prospects, one possibility is to develop a system that allows writers to check the structure of their arguments while composing their text, extending the authors' previous research. Furthermore, similar to existing studies of tabular data [16], approaches such as designing ontologies for structuring documents in different domains, such as legal judgments, or learning the ontologies themselves, can also be considered. These challenges will be addressed in future research.

The code for the results of this analysis, along with examples of the output from ChatGPT, can be found in [17]. For precise prompts, please refer to that source.

## 6. Acknowledgements

This research was supported by JST's ACT-X program (JPMJAX23CA), JST FOREST Program (JPMJFR216Q), the UTEC-UTokyo FSI Research Grant Program, and the Grant-in-Aid for Scientific Research (KAKENHI) (JP24K03043).

# 階層ナラティブ分析で紐解く生成 AI に対する認識の構造

松岡季音声 1) 松本拓樹 2) 吉田崇裕 3) 渡辺智裕 2) 近藤亮磨 2),3) 久野遼平 2),3)
1) 東京大学大学院 法学政治学研究科 2) 東京大学大学院 情報理工学系研究科 3) キヤノングローバル戦略研究所

**概要：**文書には著者の視点や物事の捉え方が反映される．そのため，文書を読み込むことは人文学や社会科学における重要な研究手法とされている．しかし，従来の感情分析やトピックモデリングといったテキストマイニング技術では，議論の構造を深く明らかにするには限界がある．この課題に対処するために，本研究では大規模言語モデルを活用し，これらの構造を抽出して階層的な枠組みに整理する手法を提案する．提案手法の検証として，日本の文化庁が収集した生成 AI に関するパブリック・コメントを分析し，生成 AI に対して肯定的な者と否定的な者のナラティブ構造を比較する．この分析により，生成 AI に対する賛否の異なる意見に影響を与える要因が明白に可視化される．

## 1．まえがき

文書には，書き手の物事に対する捉え方が反映される．そのため，文献を読み込むことは，人文科学に限らず，幅広い文系分野において，研究や情報収集の方法として確立されてきた．法学者や法曹が裁判の判決文を読み込むのは，裁判官の判断によって蓄積された規範を読み解くためである．政策当局も，有識者会議やパブリック・コメントで集められた人々の意見を統合し，それを政策原案に反映させている．特に，公文書管理法施行令やガイドラインが改正され，国の公文書が電子管理を基本とするようになった現在，より有効な公的文書の分析ツールの開発が強く求められていることは言うまでもない．このように，書き手のナラティブ構造を分析することは，広く重要なテーマである[1-5].

書き手の物事に対する捉え方の分析が面白くなるのは，書き手ごとの重要な違いを見つけたときである．それ自体は，従来の自然言語処理でも大まかには分析可能である．例えば，極性分析を用いることで肯定的な意見や否定的な意見の割合を分析することができる．トピックモデル[6]を用いることで，書き手ごとにどのような話題に注目したかをキーワードとして抽出できる．また，因果表現を含む文を抽出し，それらをつなぎ合わせることで，書き手が認知している因果構造をまとめることもできる[7,8].

もう少しナラティブ構造に踏み込んだ先行研究としては[1]にまとめられている一連の研究がある．[1]では，ナラティブを「ランダムではない，つながりのある一連の出来事」[9]として定義し，係り受け解析や共参照解析を用いて，ストーリーに参加する人物，その人物が関与するイベント，およびイベントの時間的な順序を示す時間データを構成要素として抽出している．しかし，これらの手法は，書き手のナラティブ構造の中の階層性を意識したものではなく，主題が何であるか，どのような事実を積み重ねてそれに辿り着いたかを理解することが容易ではない．また，書き手のナラティブ構造の階層性を正しく整理できなければ書き手間の認知の違いを抽出することも困難になる．

そこで本論文では，大規模言語モデル（Large Language Models，以下 LLMs）を活用し，書き手の論旨構造を階層的に集約する方法を提案する．本稿の提案手法は，階層構造のテンプレートを提供し，それに適合するように LLMs を駆使して元の文章から要素を抽出し，書き手の主張とそれを直接的に支える認識を構造化する．最終的には，文の属性情報を用いて書き手間の違いをネットワーク分析を通じて可視化するものである．複雑なタスクになるほど，LLMs に一度で問うだけではエラーが多く発生する．そのため，LLMs を利用する際には，3 段階に分けて処理を実行する．本稿のアプローチの有用性を検証するために，文化庁が発表したパブリック・コメント結果を用い，生成 AI に対して肯定的な者と否定的な者のナラティブ構造を比較する．

## 2．データセット

生成 AI は革新的である一方で社会的，倫理的問題の発生リスクも持つ．文化審議会著作権分科会法制度小委員会[10]は，行政運営の公正性と透明性を確保し，国民の権利と利益を保護する目的で，AI と著作権に関する考え方について広く意見を募集した．収集された意見と，それに対する回答者のデータ（「AI と著作権に関する考え方」のパブリック・コメント結果（個人）第 1〜3）が 2998 件公開されている[10]．本稿では，生成 AI と著作権に関する人々の意見を直接的に分析ため，このデータを活用する．

最初に全意見を肯定的および否定的な意見に分類した．図 1 は”gpt-4-turbo-2024-04-09”に各意見の極性を判定させた結果である．具体的には次のプロンプトを用いた．

> 次の文章には，生成 AI の規制に関する意見が含まれています．この文章の筆者の意見が生成 AI に対して肯定的，中立的，または否定的であるかを数値で評価してください．肯定的であれば 1，否定的であれば 0 に近づくように数値を設定してください．

図 1 とテキストに関する記述統計もまとめた表 1 にある通り，否定的な意見に偏っており肯定的な意見は少ないことがわかる他方で，肯定的に捉える者が全くいないわけでもないことが読み取れる．以降では 0.5 を規準に否定派と肯定派を分ける．これによって肯定的な意見の数は 277 となり否定的な意見

の数は 2721 となった．本データが社会における生成 AI に対する意見の極性分布を正確に捉えているかどうか，あるいは別の言い方をすれば社会全体の意見を代表する標本調査になっているかに関して，は議論の余地があるが，本稿ではこのデータを用いて生成 AI に対する肯定的な意見と否定的な意見の差を分析する．

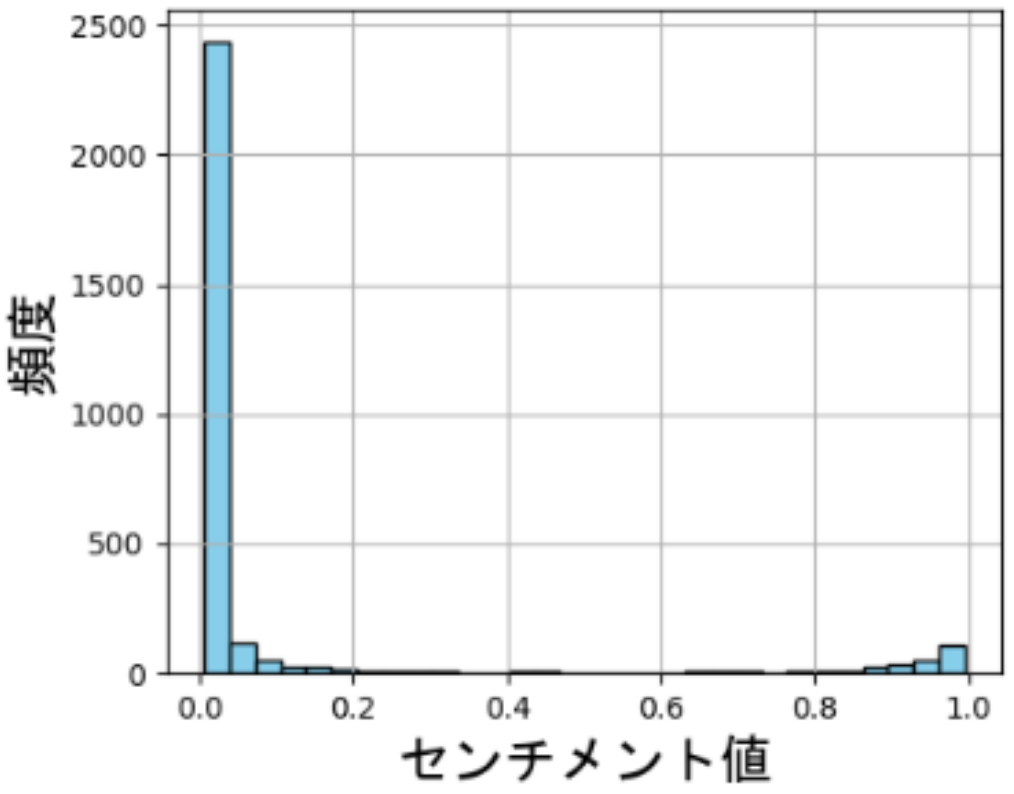

図１：極性分布

| | 平均値 | 標準偏差 | 最小値 | 中央値 | 最大値 |
|---|---|---|---|---|---|
| 文字数 | 458.4 | 438.8 | 6 | 308 | 2015 |
| 極性値 | 0.11 | 0.26 | 0.01 | 0.01 | 0.99 |

表１：データの基礎統計量

## ３．提案手法

### (1) 意見の階層構造化

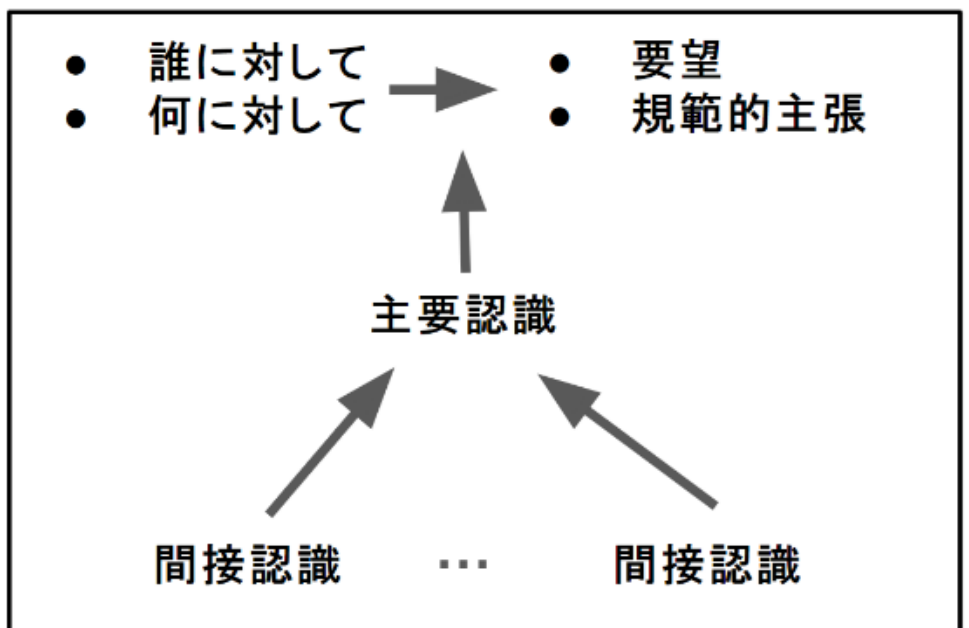

図2:意見の階層構造

一言に文章のナラティブ構造を分析するといっても，具体的に指示を出さなければ LLMs は適切に構造化できない．そこで，図 2 を意識して意見を構造化した．この階層図のポイントは 2 つである．一つは意見の中で語られている多くのものごとを書き手の認識として分類し，最終的なメインの主張である要望や規範的主張を抽出した後に，もっとも直接的につながる認識を主要認識として選び出している点である．実際の意見は冗長なものが多い．これらの中から重要な認識を抜き出すことが本提案手法の肝である．

### (2) 第 1 段階

図 2 を念頭に第 1 段階としてパブリックコメントの中の文を次の 4 パターンに分類した．

**・Pattern1: 「誰が，何に対して，どうあるべきである」という規範的主張**

**・Pattern2: 「誰が，誰に対して，どうして欲しい」という要望**

**・Pattern3: 「何が・誰が，何に対して，どういう状況を引き起こしているか」という因果関係**

**・Pattern4: 「誰が，何に対して，どう捉えているか」という認識**

である．ChatGPT に一度に聞くと本来ならば上記のパターンに分類できる場合でもいくつか文を飛ばす傾向にあることがわかった．そうした不具合が置きないように次のような工夫を加えた．

① 指示語を置き換え，言い換え表現を出力

② 意見全てをいきなり投入するのではなく 5 文ずつにわけて抽出

③ 繰り返し,「できるだけ元の表現をそのまま使うようにし，勝手に文章に書いてある表現を変化させないようにしてください」と注意したこと

④ 出力結果のエッジリストには元の文章に存在しないフレーズが含まれていないことを確認し，元の文章にないフレーズの出現回数が 0 であることを保証させたことである．

具体的には次のプロンプトを用いた．正確なものは公開コードを参照されたい[17].

次の文章には生成 AI の規制に関する意見が含まれています．以下の手順に従って解析してください．「この」,「あの」,「その」など文章内の指示語が用いられている部分が何を指しているのかを明確にして指示語の置き換えリストを出力してください．エッジリストを作成する際にはこの指示語の置き換えを参照し，それらの指示語を具体的な内容に置き換えるように注意してください．
さらに以下のパターンに基づいて情報をエッジリスト形式で抽出してください．ほとんど全ての文がこのパターンのいずれかに分類されます．そのためできるだけ全ての文を分類するようにしてください．また，できるだけ元の表現をそのまま使うようにし，勝手に文章に書いてある表現を変化させないようにしてください．Pattern1 と Pattern2 と Pattern4 の「誰が」の部分が意見を書いた人と思われる時は「筆者」としてください．
Pattern1: 「誰が，何に対して，どうあるべきである」という規範的主張
Pattern2: 「誰が，誰に対して，どうして欲しい」という要望
Pattern3: 「何が・誰が，何に対して，どういう状況を引き起こしているか」という因果関係
Pattern4: 「誰が，何に対して，どう捉えているか」という認識
…中略…
注意: Pattern4 は Pattern3 を内包する場合があります．例えば：Pattern4（誰が，（何が・誰が，何に対して，どういう状況を引き起こしているか））なので Pattern3 を内包するように書ける時はそっちを優先してください．全ての文に対して Pattern1 から Pattern4 のいずれかに該当するか

確認してください．指示語の置き換えリストに含まれている言い換え表現を，エッジリストで必ず使用するようにしてください．また，出力結果のエッジリストには元の文章に存在しないフレーズが含まれていないことを確認し，元の文章にないフレーズの出現回数が 0 であることを保証してください．

この時点での出力例は次のようなものである．意見数 2998 件に対して Pattern1 は 5637 件，Pattern2 は 1892 件，Pattern3 は 8527 件，Pattern4 は 6227 件抽出された．

"指示語の置き換え": [
"「ここでは」 = 「画像系のクリエイターに相当する仕事をしている者の懸念点」"],
"エッジリスト": [
"Pattern3 (生成 AI, クリエイターや実演家, 懸念点を引き起こしている)",
"Pattern3 (過度の AI 規制，クリエイター, AI 技術のメリットを大きく損なう危険性を引き起こしている)",
"Pattern4 (筆者, (生成 AI, クリエイターや実演家，懸念点を引き起こしている)，諸問題に対する懸念として捉えている)",
"Pattern4 (筆者, (過度の AI 規制，クリエイター, AI 技術のメリットを大きく損なう危険性を引き起こしている)，過度の規制として捉えている)",
"Pattern3 (物価高・少子化，クリエイター周辺，人的リソースの不足を引き起こしている)" ]

**(3) 第 2 段階**

Pattern1 から Pattern4 までででは図 2 に記した要望や規範的要件を分析できていない．そこで第 2 段階では要望や規範的要件と直接関係する認識を探し出し，それを支える間接認識を導くために，上記基本 4 パターンに対し，さらに下記枠組みを抽出した．

**・Pattern5: 因果関係(Pattern3)や認識 (Pattern4) が他の要素 (Pattern1 の規範的主張，Pattern2 の要望) に影響を与えている場合の関連性**

Pattern5 の抽出に関しては各意見に対して第 1 段階で抽出した Pattern1 から Pattern4 と元の文を両方入力とした上で Pattern5 を抽出するように指示を出した．第 2 段階では第 1 段階に加えて次の工夫を加えた．

⑤ 指定したエッジリストの形式よりも要素数の多いタプルなど形式を守っていないものを出力しないこと

⑥ 全ての Pattern1 と Pattern2 について，それぞれの対応関係を見つけ，Pattern1 の規範的主張や Pattern2 の要望が存在しない場合，エッジリストは空で返すことの指示を追加した．

具体的には次のプロンプトを用いた．

次に元の文章とこれまで抽出した Pattern1 から Pattern4 までの結果を元に以下の手順に従って解析してください：
まず「この」，「あの」，「その」など文章内の指示語...中略...さらに以下のパターンに基づいて情報をエッジリスト形式で抽出してください．できるだけ元の表現をそのまま使うようにし，勝手に文章に書いてある表現を変化させないようにしてください．ここで新たに Pattern1 から Pattern4 そのものを抽出する必要はなく，今回は Pattern5 だけに絞ってください．
Pattern5: 因果関係(Pattern3)や認識（Pattern4）が他の要素（Pattern1 の規範的主張, Pattern2 の要望）に影響を与えている場合の関連性
...中略...
全ての Pattern1 と Pattern2 について，以前に抽出した Pattern1 から Pattern4 までとの対応関係を検証し，Pattern5 のどの条件に該当するか確認してください．テキスト内で「誰が」の部分が意見を書いた人と思われる時は「筆者」としてください．指示語の置き換えリストに含まれている言い換え表現を，エッジリストで必ず使用するようにしてください．また，出力結果のエッジリストには元の文章に存在しないフレーズが含まれていないことを確認し，元の文章にないフレーズの出現回数が 0 であることを保証してください．さらに，指定されたエッジリストの形式を厳守し，それよりも多くの要素を持つタプルなどを出力しないでください．Pattern1 の規範的主張や Pattern2 の要望が存在しない場合は，エッジリストを空で返してください．抽出された Pattern は，原文を正確に保ち，一言一句変更せずに出力してください．最も重要なのは，元の文章を無断で変更しないことです．...以下略...

第 2 段階の出力結果は次の通りである．このステップにより Pattern5 は 6054 件抽出された．

"指示語の置き換え": [
"「この可能性」 = 「生成 AI による生産力向上の可能性」",
"「その機会」 = 「生成 AI による生産力向上の機会」"],
"エッジリスト": [
"Pattern5 (筆者, (物価, クリエイターの報酬, 相対的に低くなる状況を引き起こしている), (スタッフと工数の管理, 成果物の品質維持を求められる))",
"Pattern5 (筆者, (生成 AI 技術全体の停滞, 本国の文化活動, 大きな後退を引き起こす状況を引き起こしている), (クリエイターと社会全体, 個々人の権利保護と AI 技術の発展のバランスを取りつつ前進して欲しい))"]

**(4) 第 3 段階**

図に記した形だとそれぞれの要素の関係性を一目で把握することができず分析しにくい．そのため，再度 ChatGPT を使って, 出力データを文章化させた．具体的には，Pattern5 を「A が B であるから，C は D であるべきである」又は「A が B だから，E に対して F をして欲しい」の形に各要素を 10 文字以内でまとめるよう指示を出した．

具体的にはプロンプトを使用した．

> 次のような文章が与えられます．与えられた文章は,「文 1 であるから文 2 をしてほしい」という要望か,「文 1 であるから文 2 であるべきである」という規範的主張を表しています．
> この文章は，文 2 の内容に応じて二つの形式で要約されます：
> 文 2 が規範的主張をしている場合，要約は「A が B であるから，C は D であるべきである」となります．この要約の作成にあたって，文 1 からは「A が B である」という部分を抽出し，文 2 からは「C は D であるべきである」または「C は D すべきである」という表現を抽出します．
> 文 2 が要望を表している場合，要約は「A が B だから，E に対して F をして欲しい」となります．この要約の作成にあたって，文 1 からは「A が B である」という部分を抽出し，文 2 からは「E に対して F をして欲しい」という表現を抽出します．
> A から F までのそれぞれの項目は 10 文字以内である必要があり，接続詞や助詞は文字数に含めません．最終的な要約も含めて示してください．
> 文 2 が規範的主張をしている場合は，E と F は「NA」であるとし，逆に文 2 が要望を表している場合は，C と D は「NA」であるとします．
> エッジリストには元の文章に存在しないフレーズが含まれていないことを確認し，元の文章にないフレーズの出現回数が 0 である...以下略...

出力結果を表形式でまとめると表 2 のようになる．

| type | A | B | C | D | E | F |
|---|---|---|---|---|---|---|
| 規範的主張 | 画像生成AIの許諾不要な状態 | 作品の作成者、著作権者ならびに著作者の利益と機会に対して平等を著しく毀損している | 画像生成AIの許諾不要な状態 | 許容しないこと | NaN | NaN |
| 要望 | 生成AI | 日本の経済に対して低迷 | NaN | NaN | 日本政府 | 日本を助ける |

表 2：第 3 段階後の結果

**(5) ネットワーク可視化**

このようにして抽出した A から F までの各出力結果を，Universal Sentence Encoder[11]でベクトル埋め込みに変換し，次に DBSCAN[12]を用いクラスタリングをすることで，同じ意味内容のものを一つにまとめた．これらのクラスタ番号を A_c，B_c，C_c，D_c，E_c，F_c とした．各クラスタの代表表現はそのクラスタの真ん中にあるものを選んだ．

その上で A_c→B_c，B_c→C_c，C_c→D_c と A_c→B_c，B_c→E_c，E_c→F_c を肯定的な意見と否定的な意見に分けて繋ぎ，有向ネットワークとして可視化した．つまり，本ネットワークにおいてエッジに対応しているものは認識（「〇が□であるから」），規範的主張（「〇は□であるべき」），要望（「〇に対して□をして欲しい」）の 3 種類である．ネットワーク全体としては次数二以上のエッジにのみ絞り，ポジティブなものに関しては件数が少ないことを考慮し次数が 1 でもよいものとした．また，ノードに位置に関しては，同じものであれば，肯定派であっても，否定派であっても同じ位置に描画されるように工夫した．位置は Gephi[13]に搭載されている ForceAtals2 を用いた[14]．各ノードの色はモジュラリティ最大化によって得たコミュニティである[15]．このコミュニティについても肯定はと否定派のネットワークの和集合を用いて計算した．

## ４．結果

このようにして得た否定派のネットワークと肯定派のネットワークが，それぞれ図 4 と図 5 である．前述の通り，どちらのネットワークにおいても共通するノードは同じ位置に表示している．その情報を参考に，大まかな対応位置関係を示したのが(1)から(7)までの数字である．この位置情報ごとにそれぞれの特徴を考察する．

まず，位置(1)において軸となるノードは「生成 AI の生成物」である．否定派は，(1)の近くにおいて「AI の扱い」について「過度に制限すべきではない」と一部あるものの,「生成 AI の生成物」に対して「許容するべきではない」,「脅迫をしている」,「クリエイターに利益なし」,「問題を引き起こす」,「窃盗」,「創作文化に対して破壊を引き起こしている」など強固に反対する姿勢が強く認められる．こうした意見が至る所に散見されるため本ネットワークにおいては「生成 AI の生成物」が最大次数のノードになっている．それに対して肯定派は「生成 AI の研究」や「生成 AI に対する著作物の学習および使用」に対して「過度に制限すべきではない」と主張しており,「AI が無断で学習」することを「クリエイターへの敬意に欠けている」としつつも「生成 AI の生成物」に対しては「止めるべきではない」や「著作物性を認めるべき」など寛容な意見が目立つ．つまり，否定派のように強固に反対している意見が見られない．

次に，(2)において肯定派と否定派の認知の分かれ目となっているノードは「AI が無断で学習」で

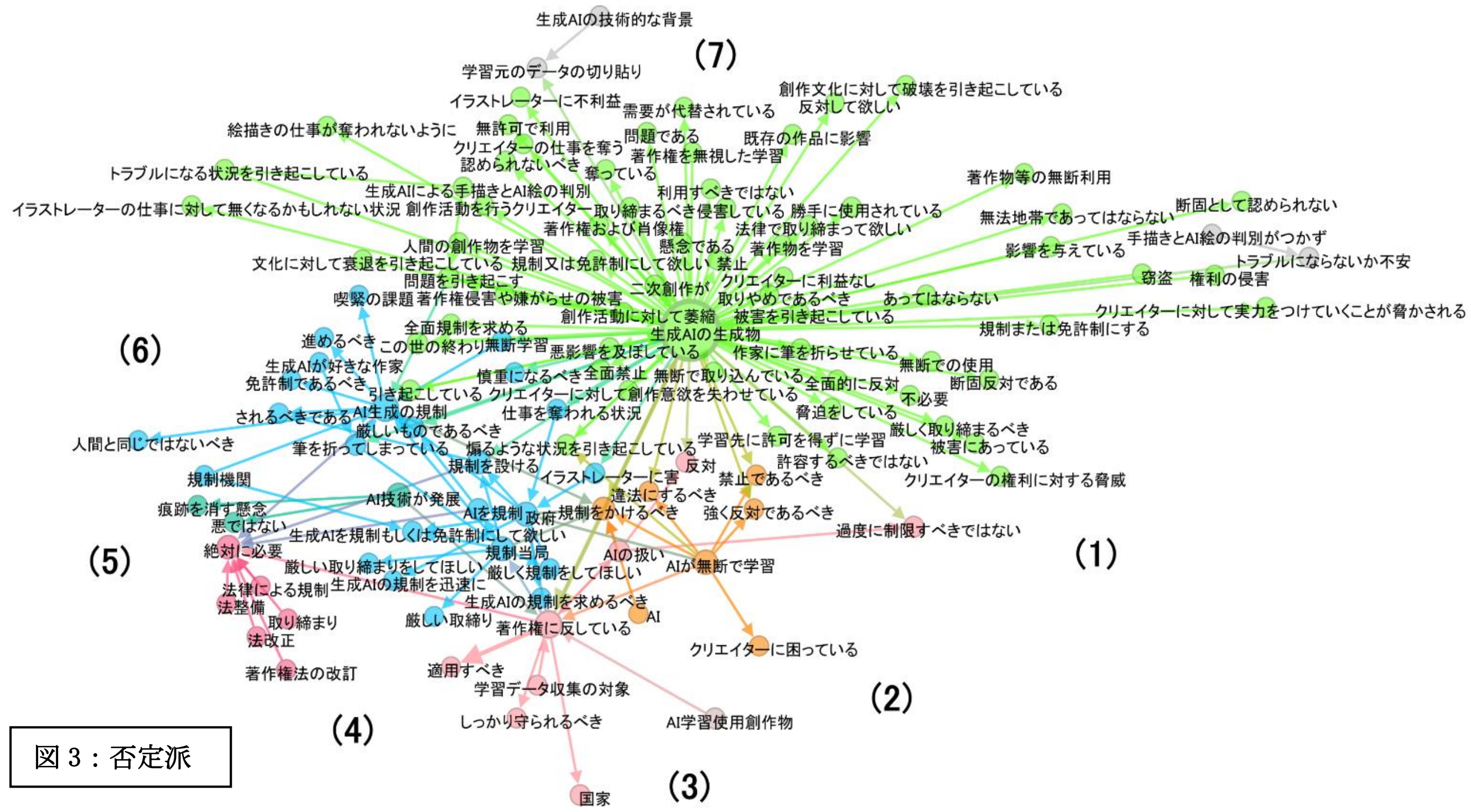

図３：否定派

ある．否定派は「AI が無断で学習」することに対して，「クリエイターに困っている」や「強く反対であるべき」といった否定的な意見が多いのに対し，肯定派は「AI が無断で学習」することに対して「全面的に認められるべき」という意見や，「著作権での制限」に関しても「コンピューターサイエンス」の「発展を妨げないように」など，新しい技術に対して肯定的な意見が目立つ．

(3)において軸となるノードは「AI 学習使用創作物」である．否定派が「AI 学習使用創作物」に対して「著作権に反している」と厳しく批判しているのに対し，肯定派は「AI 学習使用創作物」に対して「訓練データの利用」に関して「制限」を「設けるべきではない」と，クリエイターというよりはエンジニア側の視点に立った意見を述べていることは注目に値する．パブリック・コメント自体には，書き手が述べない限り属性情報がわからないが，ナラティブ構造を見るだけで属性がある程度想像できることは興味深い．

(4)と(5)に関しては規制に関する話題が多い． (4)にある「規制当局」に対して否定派が「著作権に反している」から「規制当局」に対して「AI を規制」して欲しい「生成 AI を規制もしくは免許制にして欲しい」「厳しい取り締まりをしてほしい」と規制強化の意見を強く述べている．仮に免許制を特許制と捉えるならばそれは本質的に国民が当該事業を自由に行う権利を本来的に有さず，国が独占するものであるため，それは相当強い意見であると見ることができる．それに対して肯定派は「規制当局」に対して「具体的な例を出して周知」して欲しいと要望を述べるなど生成 AI ユーザー側にたっている．

また，(5)の「AI 技術が発展」に対して肯定派は「産業発展にほぼ同義」であるとか近くに「諸外国の推進」が「技術格差による危険性」につながるなど経済発展に関連させた議論が目立つ．また，興味深い意見としては，「AI 技術が発展」が「創作活動の萎縮」を生み出さないようにクリエイター側も「準備運動」が「絶対必要」など技術発展をクリエイター側も考慮すべきもあった．それに対して否定派は「AI 技術が発展」に対して「痕跡を消す懸念」につながっている．これは元の意見も参照した所，著作権法の依拠性の根拠となる痕跡を消せる可能性について言及しており，その点でさらなる AI 技術の発展によって現行法制度が形骸化する懸念を示したものである．

(6)も同様に規制に関する話題であるが，肯定派が「AI 生成の規制」に対して「現状の法規制にとどめる」や「営利利用に規制」するなど，現状維持の意見が目立つのに対し，否定派は「AI 生成の規制」に対して「免許制であるべき」「厳しいものであるべき」「人間と同じではないべき」など，積極的に法制度を改正しようとする姿勢が興味深い．日本の著作権法は，2018 年から 2021 年にかけて 4 回改正されてきた．特に，2018 年の改正では，情報解析のための著作物の利用について，従来より柔軟な権利制限を規定することで，AI の学習が容易になったという背景がある（第 30 条の 4，第 47 条の 4，第 47 条の 5）．これらを踏まえた結果かは不明だが，肯定派は機械学習応用に対して一定の理解を示した法制度に対し，現状維持を望んでいる可能性がある．

最後に(7)であるが，ここで否定派は「生成 AI の生成物」に対して，（1）と同様に「創作文化に対して破壊を引き起こしている」「窃盗」「クリエイターに対して実力をつけていくことが脅かされる」など，強固に反対している意見が目立つ．むしろ，最大次数ノードである「生成 AI の生成物」に対し

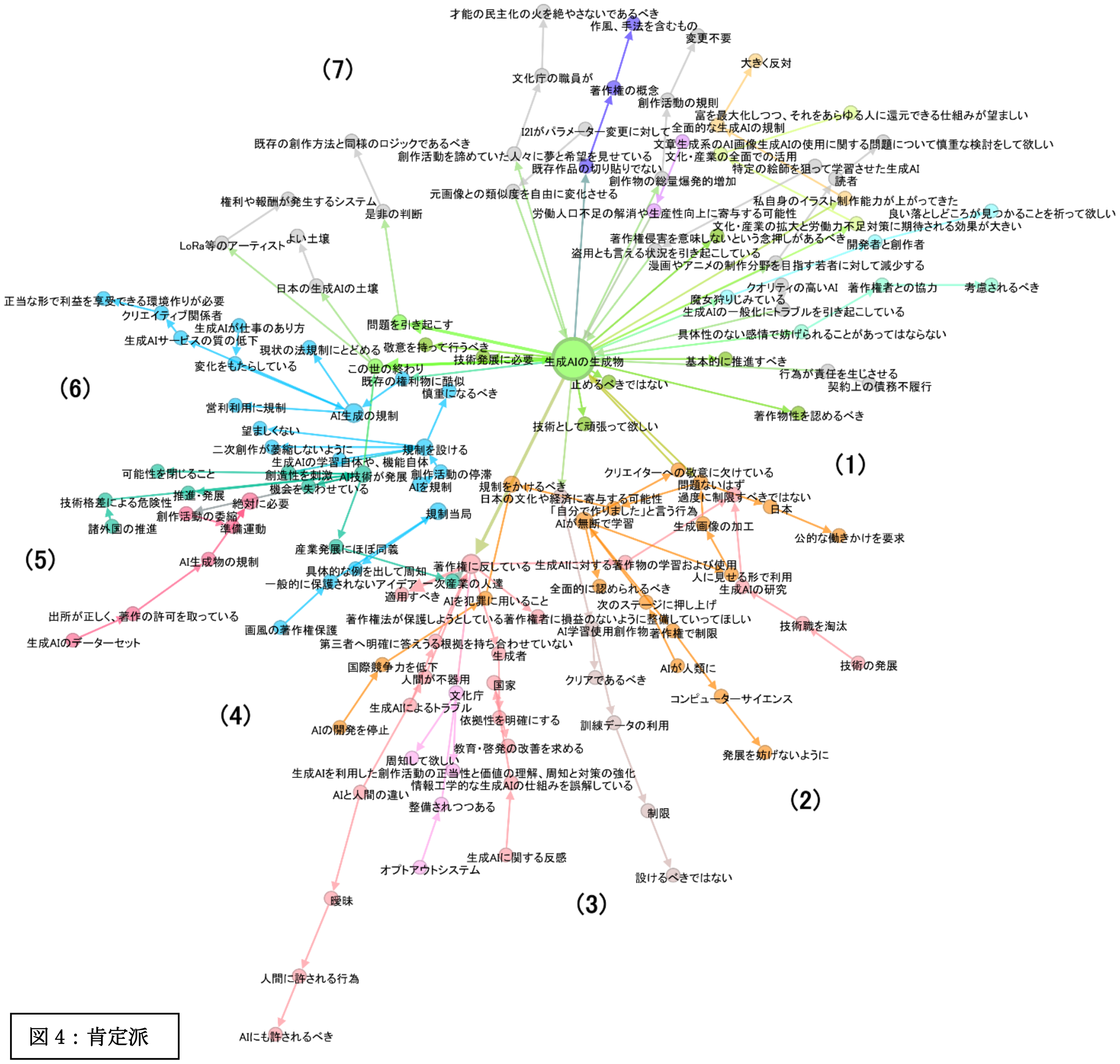

図 4：肯定派

て，直接的に批判的な思いがこの位置にまで広がっている．それに対して肯定派は「生成 AI の生成物」に対して，「才能の民主化の火を絶やさないであるべき」「富を最大化しつつ，それをあらゆる人に還元できる仕組みが望ましい」「生成 AI の仕様に関する問題について慎重な検討をして欲しい」「創作活動を諦めていた人々に夢と希望を見せている」「良い落としどころが見つかることを祈って欲しい」など，新しい技術を有効活用しようという期待が述べられている．

このように，意見の階層構造を意識してパブリック・コメントから主要な主張を抽出することで，生成 AI に対する否定派と肯定派のナラティブ構造の違いを比較・分析することができた．見やすいように，最後に両者の違いをまとめたものを表 3 に掲載する．

## 5．最後に

本論文では，LLMs を活用し，書き手の論旨構造を階層的に整理するとともに，文の属性情報を用いて話者間の違いをネットワーク分析によって可視化する方法を提案した．提案手法は，人や関係性，時といった物語としてのナラティブ構造[1]とは異なり，パブリック・コメントや判決文のように意見構造が明確に存在する文書に適していると考えられる．そのため，提案手法の有効性を示すために，文化庁が発表した生成 AI に対するパブリック・コメントの結果を用いて分析を行った．まず，各意見に対して肯定的な意見と否定的な意見を，意見全体の極性を用いて判定し，肯定的な意見と否定的な意見の中ののナラティブ構造の違いに焦点を当てた分析を実施した．その結果，様々な論点において，従来手法と比較して，意見の違いが生じる要因をより明確に可視化できた．

今後の展望としては，筆者らの過去研究を拡張する形で，書き手が文章を執筆している最中に自分の意見構造を確認できるシステムを構築することが考

| 位置 | 軸となるノード | 肯定派 | 否定派 |
|---|---|---|---|
| (1) | 「生成AIの生成物」「過度に制限すべきではない」 | 「生成AIの研究」に対して「過度に制限すべきではない」という意見を軸に「日本」に対して「公的な働きかけ」を要求している点や「AIが無断で学習」することを「クリエーターへの敬意に欠けている」としつつも「生成AIの生成物」に対しては「止めるべきではない」や「著作物性を認めるべき」など寛容な意見が目立つ。否定派のように強固に反対している意見が見られない。 | 「AIの扱い」について「過度に制限すべきではない」とあるものの「生成AIの生成物」に対して「許容するべきではない」や「脅迫をしている」など強固に反対している意見が目立つ。 |
| (2) | 「AIが無断で学習」 | 「AIが無断で学習」することに対して「全面的に認められるべき」とあったり「著作権での制限」に関しても「コンピューターサイエンス」の「発展を妨げないように」など新しい技術に対して肯定的な意見が目立つ。 | 「AIが無断で学習」することに対して「クリエイターに困っている」という意見や「強く反対であるべき」などの否定的意見が多い。 |
| (3) | 「AI学習使用創作物」 | 「AI学習使用創作物」に対して「訓練データの利用」に関しては「制限」を「設けるべきではない」とクリエーターというよりはエンジニア側の視点にたった意見を述べている。 | 「AI学習使用創作物」に対して「著作権に反している」とばっさり切り捨てている。 |
| (4) | 「規制当局」 | 「規制当局」に対して「具体的な例を出して周知」して欲しいと要望を述べるなど生成AIユーザー側にたっている。 | 「規制当局」に対して「AIを規制」して欲しいとか「生成AIを規制もしくは免許制にして欲しい」と規制強化の意見を述べている。 |
| (5) | 「AI技術が発展」 | 「AI技術が発展」に対して「産業発展にほぼ同義」であるとか近くに「諸外国の推進」が「技術格差による危険性」につながるなど経済発展に関連させた議論が目立つ。 | 「AI技術が発展」に対して「痕跡を消す懸念」につながっている。これは元の意見も参照した所、著作権法の依拠性の根拠となる痕跡を消せてしまう可能性について言及しており、その点でさらなるAI技術の発展によって現行法制度が形骸化するのではないかという懸念を示したものであった。 |
| (6) | 「AI生成の規制」 | 「AI生成の規制」に対しては「現状の法規制にとどめる」や「営利利用に規制」するなど現状維持の意見が目立つ。 | 「AI生成の規制」に対して「免許制であるべき」「厳しいものであるべき」「人間と同じではないべき」などの積極的に規制する方向の意見が目立つ。 |
| (7) | 「生成AIの生成物」 | 「才能の民主化の火を絶やさないであるべき」「富を最大化しつつ、それをあらゆる人に還元できる仕組みが望ましい」「生成AIの仕様に関する問題について慎重な検討をして欲しい」「創作活動を諦めていた人々に夢と希望を見せている」「良い落としどころが見つかることを祈って欲しい」など単発の要望や考察が並んでいる。 | （１）と同様に「創作文化に対して破壊を引き起こしている」「窃盗」「クリエイターに対して実力をつけていくことが脅かされる」など強固に反対している意見が目立つ。 |

表３：肯定派と否定派の比較

えられる．また，表形式のデータに関しては既存研究があるように[16]，判決文など異なるドメインにおいても，文書を構造化するためのオントロジーを設計したり，オントロジー自体を学習したりするアプローチも考えられる．それらの課題は今後の研究テーマとして取り組む予定である．

本分析の結果のコードは[17]に結果例と ChatGPT による出力結果も共に掲載してある．特に正確なプロンプトに関してはそちらを参照されたい．

## ６．謝辞

本研究は JST の ACT-X（JPMJAX23CA），JST の創発的研究支援事業（JPMJFR216Q），UTEC-UTokyo FSI 研究助成プログラム並びに科研費（JP24K03043）による支援を受けた．

## 参考文献